\documentclass[11pt]{article}

\usepackage[margin=1in]{geometry}
\usepackage{setspace}
\usepackage{graphicx}
\usepackage{booktabs}
\usepackage{hyperref}
\usepackage{natbib}
\usepackage{graphicx}
\usepackage{amsmath}
\usepackage{amssymb}
\usepackage{color}

\title{Input Order Shapes LLM Semantic Alignment in Multi-Document Summarization}
\author{
  Jing Ma \\
  University of Zurich \\}

\date{}

\begin{document}
\maketitle
\begin{abstract}
Large language models (LLMs) are now used in settings such as Google's \textit{AI Overviews}, where it summarizes multiple long documents. However, it remains unclear whether they weight all inputs equally. Focusing on abortion-related news, we construct 40 pro-neutral-con article triplets, permute each triplet into six input orders, and prompt Gemini 2.5 Flash to generate a neutral overview. We evaluate each summary against its source articles using ROUGE-L (lexical overlap), BERTScore (semantic similarity), and SummaC (factual consistency). One-way ANOVA reveals a significant primacy effect for BERTScore across all stances, indicating that summaries are more semantically aligned with the first-seen article. Pairwise comparisons further show that Position~1 differs significantly from Positions~2 and~3, while the latter two do not differ from each other, confirming a selective preference for the first document. The findings present risks for applications that rely on LLM-generated overviews and for agentic AI systems, where the steps involving LLMs can disproportionately influence downstream actions.

\end{abstract}

\section{Introduction}
The rapid evolution of generative AI, which serves as a core subcomponent in agentic systems, has accelerated the development and early deployment of agentic AI systems across diverse domains. Agentic AI is characterized by its ability to pursue complex objectives adaptively with limited direct supervision, exhibiting a high degree of operational autonomy \citep{mukherjee2025agentic,shavit2023practices}. A common mechanism enabling this autonomy is \textit{goal decomposition}, wherein a high-level user objective is parsed into a sequence of manageable subtasks \citep{Sapkota2025AIAV}. To execute these subtasks effectively, such systems typically engage in multi-step reasoning and planning, retrieving and interpreting external information (e.g., via retrieval-augmented generation and tool use) to guide subsequent actions \citep{Sapkota2025AIAV,shavit2023practices}. Consequently, the reliability of an agentic system is intrinsically linked to how accurately its underlying Large Language Models (LLMs) retrieve, prioritize, and synthesize information from instructions and external documents.

A popular implementation of this retrieval-and-synthesis capability is the \textit{AI Overviews} feature within Google Search. Powered by the Gemini model family, the system operates as a retrieval-enhanced generation pipeline, processing multiple long-context documents to produce concise, cited summaries \citep{reid2024generative}. While this functionality improves user efficiency by reducing the need to manually synthesize information across diverse sources, it raises critical questions regarding information reliability. In particular, when an LLM integrates multiple documents into a single summary, it becomes essential to assess whether the model weighs all input sources uniformly or whether it exhibits systematic structural biases.

Prior research has examined the ideological biases embedded in LLMs. For example, \citet{Santurkar2023WhoseOD} used survey-style prompts to show that models such as GPT-4 and Claude frequently express liberal-leaning social values. Within the domain of news summarization, \citet{Vijay2024WhenNS} proposed quantitative measures such as vocabulary divergence (VD), stance polarization index (SPI), and algorithmic monoculture index (AMI) to demonstrate that LLMs often converge on a narrow ideological framing. While previous work has extensively examined ideological and content-level biases in LLM-generated summaries, few studies address whether LLMs exhibit structural biases in how they weight multiple input documents. In multi-document settings, particularly those involving long concatenated inputs, it remains unknown whether LLMs prioritize some documents over others due to their position within the context window. Positional effects such as primacy bias or lost-in-the-middle have been observed in other long-context tasks, yet their implications for multi-document summarization have limited evaluation. Accordingly, this paper fills this gap by examining whether the sequential ordering of source articles influences their representational weight in LLM-generated summaries, using the topic of abortion news articles as a test case. We address the following research questions:

\vspace{1em} 

\noindent \textbf{RQ1:} Which input documents receive greater representational weight in the model’s generated summary?

\noindent \textbf{RQ2:} Does the sequential order of the input articles significantly influence the content and focus of the model's output?


\section{Related Work}
\textbf{Position Bias.}
Despite the advanced capabilities of LLMs in Natural Language Processing (NLP), recent research has shown that they are susceptible to the structural arrangement of input data, a phenomenon often referred to as position bias. Position bias describes the tendency of models to disproportionately prioritize information appearing at specific positions within the input sequence, leading to the neglect of important details elsewhere \citep{jung-etal-2019-earlier, Wang2023IsCA, Zheng2023JudgingLW, Chhabra2024RevisitingZA}. In the context of multi-document summarization (MDS), where models operate over multiple long documents concatenated into a single context, this bias implies that the likelihood of information being incorporated into the summary can depend on a document’s placement within the context window. Consequently, LLM-generated summaries may exhibit skewed perspectives or imbalanced content coverage depending on the input order. 

A manifestation of this issue is the Lost in the Middle phenomenon. \citet{Liu2023LostIT} evaluated LLM performance on multi-document question answering and key–value retrieval, observing a characteristic U-shaped performance curve: accuracy peaks when relevant information appears at the beginning (primacy) or end (recency) of the context but degrades when the information is placed in the middle. While their analysis centered on tasks with short, limited-token outputs, related work by \citet{Koo2023BenchmarkingCB} demonstrates that similar structural biases emerge in annotation settings. When LLMs are used as evaluators, they exhibit a strong positional bias, systematically favoring the first or last option presented, even when the highest-quality response appears elsewhere in the sequence. 

More specifically, position bias often interpreted as lead bias, where the initial segments of the input receive disproportionate emphasis. \citet{Chhabra2024RevisitingZA} found that while some models (e.g., GPT-3.5-Turbo) exhibit relatively mild lead bias, others display substantial sensitivity to early-positioned content, especially in extreme summarization settings. Beyond degrading performance, such positional sensitivity raises pressing fairness concerns. \citet{Olabisi2024UnderstandingPB} demonstrate that in social multi-document summarization, ordering tweets by dialect can cause summaries to favor whichever group appears first. This indicates that input ordering can profoundly affect the representational fairness of the resulting summary, even when the generated text remains fluent and well-formed.

\textbf{Evaluation Metrics.} Evaluating the quality of generated summaries requires a multi-dimensional approach that considers lexical overlap, semantic similarity, and factual consistency. Traditional metrics such as BLEU \citep{papineni-etal-2002-bleu} and ROUGE-L \citep{lin-2004-rouge} assess quality through n-gram overlap between system outputs and reference summaries. Although widely adopted, these metrics often fail to handle paraphrasing, where two texts convey the same meaning with different wording. To overcome these limitations, \citet{Zhang2019BERTScoreET} proposed BERTScore, which embeds tokens using pre-trained contextual encoders and computes similarity via cosine distance. By capturing semantic relationships and paraphrastic variation, BERTScore offers a better measure of summary quality in settings requiring concise abstraction.

Beyond semantic similarity, it is essential to assess whether a summary is factually consistent with its source documents. Several metrics have been developed for this purpose. FactCC \citep{kryscinski-etal-2020-evaluating} employs a trained classifier to determine whether each summary sentence is supported by the source text. However, subsequent analyses show that FactCC can be unstable, occasionally misclassifying sentences that are supported by the original text. BLANC \citep{vasilyev-etal-2020-fill}, a reference-free metric, evaluates how effectively a summary helps a LM reconstruct masked portions of the source document. Yet BLANC scores are sensitive to the choice of underlying LM and sometimes correlate more with lexical overlap than with true semantic entailment.
To address these limitations, SummaC \citep{Laban2021SummaCRN} evaluates factual consistency through a Natural Language Inference (NLI) framework, typically using models such as RoBERTa-MNLI. SummaC operates at the sentence level, assessing whether the source text support each statement in the summary. Empirical evaluations show that SummaC generally outperforms earlier metrics such as FactCC.

To ensure a comprehensive assessment in this study, we adopt a multiple evaluation strategy: ROUGE-L to capture lexical overlap, BERTScore to measure semantic similarity, and SummaC to evaluate factual consistency between the generated summaries and the original articles.
\section{Methods}

\begin{figure}[!ht]
    \centering
    \includegraphics[width=1\linewidth]{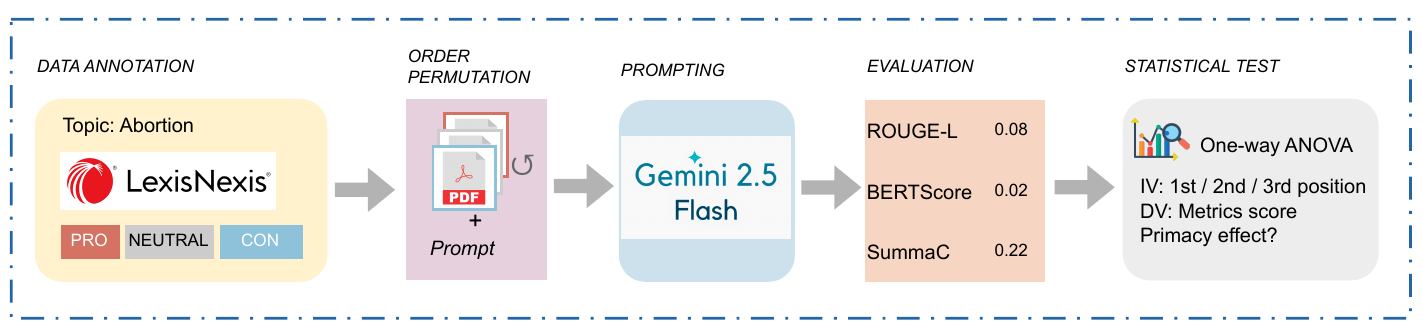}
    \caption{Overview of the experimental pipeline. Triplets of PRO/NEUTRAL/CON news articles on abortion were collected and annotated, permuted into six input orders, and summarized by Gemini 2.5 Flash. Resulting summaries were evaluated using ROUGE-L, BERTScore, and SummaC, and statistical differences across positions were tested with one-way ANOVA to detect primacy effects.}
    \label{fig:pipeline}
\end{figure}

Figure~\ref{fig:pipeline} illustrates the pipeline employed in this study, spanning from data acquisition to statistical test. 
The process starts with the data collection phase, where news articles related to the controversial topic of abortion were sourced via LexisNexis. These documents were retrieved in PDF format. Subsequently, we categorized the positional stance of each article as supportive (pro), neutral and  conservative (con) through mannual annotation. 
Following the generation of summaries based on varying input permutations, we quantified the similarity between the generated summary and the original source articles with three distinct metrics: ROUGE-L (lexical overlap), BERTScore (semantic similarity), and SummaC (factual consistency). Finally, to evaluate whether the sequence of input documents significantly influences the resulting summary, we conducted a one-way Analysis of Variance (ANOVA).

\subsection{Data Collection}

\textbf{Search Strategy.} We collected articles related to topic of abortion from LexisNexis. To ensure comprehensive coverage, we constructed a Boolean search query utilizing the \texttt{OR} operator to retrieve articles containing at least one of the following terms: ``abortion,'' ``reproductive rights,'' ``Roe v Wade,'' ``Dobbs decision,'' ``pro-life,'' ``pro-choice,'' ``anti-abortion,'' or ``right to choose.'' We also restricted news sources to six major outlets: \textit{CNN Wire}, \textit{USA Today}, \textit{USA Today Online}, \textit{The New York Times}, \textit{TheHill.com} and \textit{The Associated Press} to capture a representative cross-section of mainstream media perspectives.

\textbf{Annotation and Pre-processing.} We built a corpus of 120 articles and manually annotated each for positional stance (PRO, NEUTRAL, or CON) based mainly on its textual content and framing of abortion rights. A PRO stance denotes explicit or implicit support for abortion rights, whereas a CON stance indicates opposition. Articles presenting factual information devoid of clear evaluative framing were labeled NEUTRAL. Crucially, these annotations were derived exclusively from the provided text, independent of the publication's known ideological leaning or reputation.

\textbf{Dataset Construction.} The 120 annotated articles were divided into 40 triplets. Each triplet comprised one article from each stance (CON, NEUTRAL, PRO). To control for length bias, articles within a triplet were matched by word count category (e.g., the ``500-word'' category contained articles between 500–600 words). The total word count for articles in the corpus ranged from 300 to 1,600 words. This sampling procedure ensured a balanced length distribution across stances, thereby mitigating potential confounding variables related to text length in the final analysis.

\begin{figure}[!ht]
    \centering
    \includegraphics[width=0.8\linewidth]{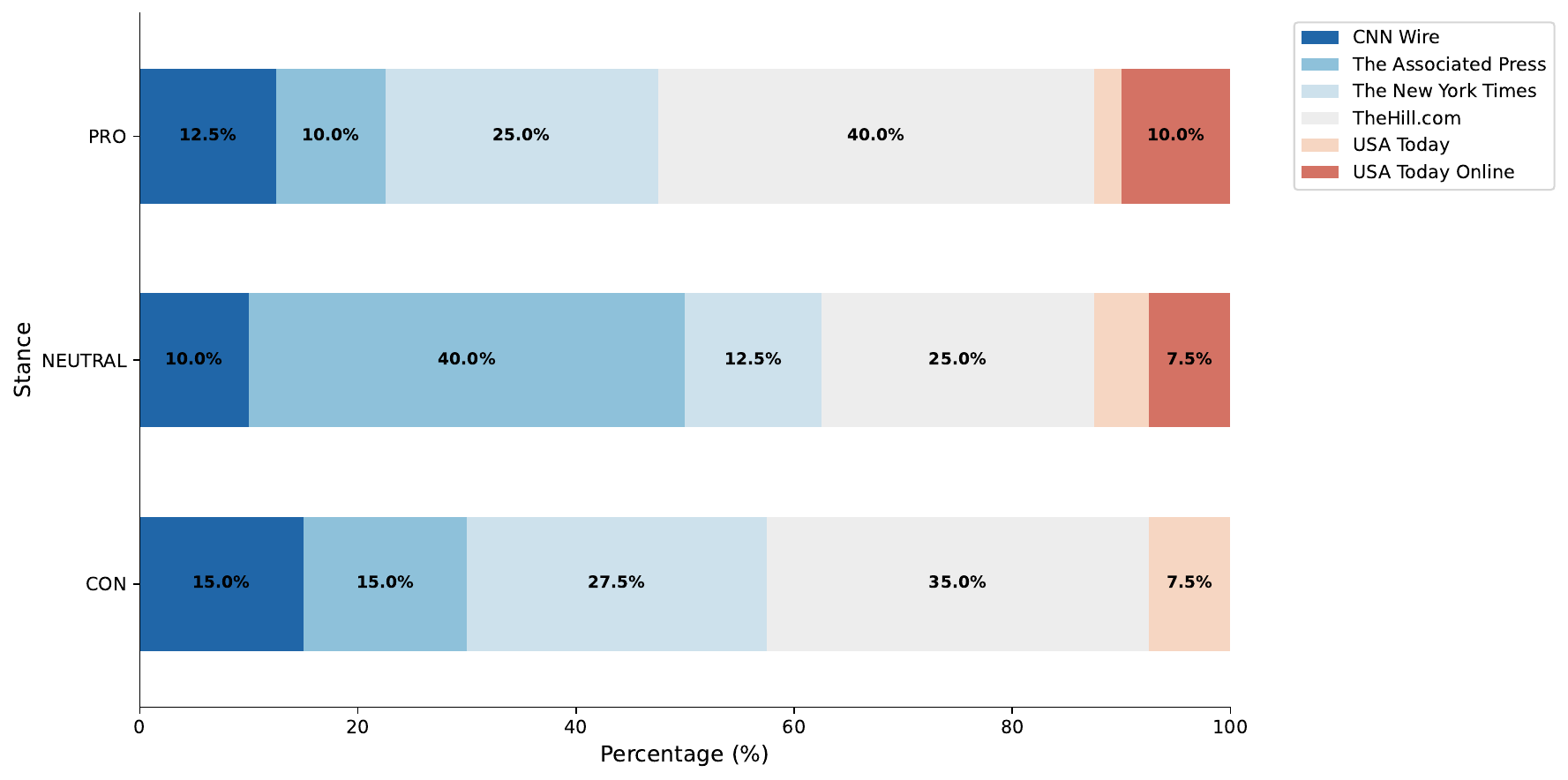}
    \caption{Distribution of six major U.S. news sources across PRO, NEUTRAL, and CON abortion stances. Each bar shows the percentage contribution of each source within a stance category.}
    \label{fig:source_distribution}
\end{figure}
Figure~\ref{fig:source_distribution} illustrates the distribution of news sources divided by stance within the collected corpus. Across all categories, \textit{TheHill.com} and \textit{The Associated Press} emerge as the predominant contributors. specifically, the CON stance is primarily represented by \textit{TheHill.com} (35\%) and \textit{The New York Times} (27.5\%). The NEUTRAL category is dominated by \textit{The Associated Press} (40\%) and \textit{TheHill.com} (25\%). Finally, the PRO stance relies heavily on \textit{TheHill.com} (40\%), followed by \textit{The New York Times} (25\%) and \textit{The Associated Press} (10\%). Overall, the dataset reflects a mix of mainstream U.S. news outlets that cover abortion from diverse political perspectives.

\subsection{Experimental Design}
To evaluate the impact of input order on summarization quality, we utilized the article triplets constructed in the data collection phase. For each triplet, we instructed the \textit{Gemini-2.5-Flash} model to generate a summary using the following prompt:
\begin{quote}
\small
\textit{``Please read three news articles on the same topic with different stances (pro, neutral, and con), and write one concise, factual, and neutral overview (150–220~words) that fairly represents all sides without speculation, personal opinions, or emotional expressions.''}
\end{quote}
The three articles were permuted into all six input orders:
\begin{align*}
    &(\text{pro}, \text{neutral}, \text{con}),\;
     (\text{pro}, \text{con}, \text{neutral}),\;
     (\text{neutral}, \text{pro}, \text{con}),\\
    &(\text{neutral}, \text{con}, \text{pro}),\;
     (\text{con}, \text{pro}, \text{neutral}),\;
     (\text{con}, \text{neutral}, \text{pro}).
\end{align*}
For each permutation, we saved the model's output in plain-text format for evaluation.

\subsection{Evaluation Procedures}
In this section, we describe how each evaluation metric is computed and what its outputs represent, covering lexical overlap, semantic similarity, and factual consistency. 

\textbf{ROUGE-L} \citep{lin-2004-rouge} measures the longest common subsequence (LCS) between a model-generated summary (hypothesis) and a reference. ROUGE-L captures sentence-level structural similarity and is therefore less sensitive to word-order variations. It is typically used as a reference-based metric, computed between a human-written reference summary and the system output. In this study, we are not concerned with the absolute ROUGE-L score; rather, we examine relative differences across input-order conditions to test for primacy effects. Because ROUGE-L reflects surface-level lexical overlap, we complement it with two additional metrics, BERTScore and SummaC, to evaluate semantic similarity and factual consistency, respectively.

To move beyond surface-level word overlap, we adopt contextualized embeddings to capture semantic similarity.
\textbf{BERTScore }\citep{Zhang2019BERTScoreET} computes token-level similarity using contextual embeddings from a pre-trained encoder (RoBERTa-large in this study). For each token in the hypothesis, BERTScore identifies the most similar token in the reference and computes precision, recall, and F1 based on these token alignments. Because contextual embeddings encode synonymy and paraphrastic variation, BERTScore is well suited for evaluating semantic similarity beyond exact lexical matches. Consistent with our study design, we focus on relative differences in BERTScore across input-order conditions, rather than the absolute similarity values, which may be low due to the abstractive nature of the task.

\textbf{SummaC} evaluates whether the generated summary is entailed by the original documents, thereby assessing factual consistency. The metric has two variants: SummaC-ZS (zero-shot, parameter-free) and SummaC-Conv (trained convolutional model). Due to its superior performance, we employ SummaC-Conv. Because the underlying RoBERTa-based NLI model supports a maximum input length of 512 tokens, we split each article into multiple chunks to ensure that every (chunk + summary) pair fits within this limit. The summary is compared against each chunk, and the final SummaC score is obtained by averaging entailment scores across all chunks, yielding a value in the range [0,1].
\section{Results}
\begin{figure}[h]
    \centering
    \includegraphics[width=\textwidth]{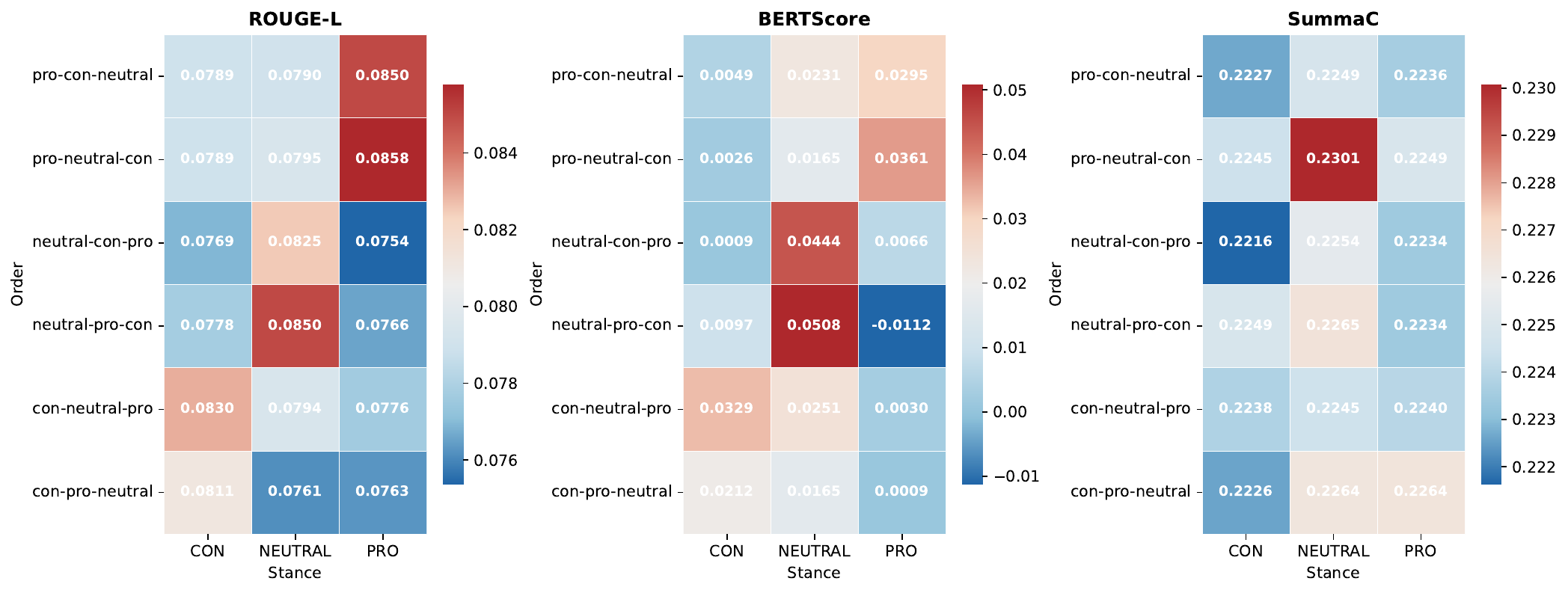}
    \caption{Effects of input order on similarity scores across ROUGE-L, BERTScore, and SummaC.
Each heatmap shows the mean metric value for a given stance (CON, NEUTRAL, PRO) under six possible input sequences.}
    \label{fig:heatmaps}
\end{figure}

After visualizing the results with heatmaps and conducting a one-way ANOVA to test whether input position significantly influences the generated summaries, we obtained the following findings. Figure~\ref{fig:heatmaps} presents the mean values of the three evaluation metrics across the six input orders and three stance categories. The mean for each condition is computed by averaging the metric scores across all article triplets with the same input order and stance. The variation in shading and metric values across different orders provides empirical evidence that input positioning shapes similarity patterns in LLM-generated summaries. In particular, summaries exhibit higher ROUGE-L and BERTScore values when compared against the document placed first in the sequence.

\begin{table}[h]
    \centering
    \begin{tabular}{l l c c c}
    \toprule
    Metric & Stance & $F$ & $p_{\text{corr}}$ & $\eta^2$ \\
    \midrule
    \textbf{BERTScore} 
        & PRO     & \textbf{9.25}  & \boldmath{$<.001^{***}$} & \textbf{0.072} \\
        & NEUTRAL & \textbf{9.06}  & \boldmath{$<.001^{***}$} & \textbf{0.071} \\
        & CON     & \textbf{4.70}  & \textbf{0.030$^{*}$}   & \textbf{0.038} \\
    \midrule
    \textbf{ROUGE-L} 
        & PRO     & 3.76  & 0.055         & 0.031 \\
        & NEUTRAL & 1.38  & 0.381         & 0.012 \\
        & CON     & 0.67  & 0.658         & 0.006 \\
    \midrule
    \textbf{SummaC}
        & PRO     & 0.28 & 0.753 & 0.002 \\
        & NEUTRAL & 0.43 & 0.732 & 0.004 \\
        & CON     & 2.34 & 0.177 & 0.019 \\
    \bottomrule
    \end{tabular}
    \caption{One-way ANOVA results across three similarity metrics. 
    $p_{\text{corr}}$ corrected using FDR-BH. 
    Significance codes: * $p<.05$, ** $p<.01$, *** $p<.001$.}
    \label{tab:anova}
\end{table}

\begin{table}[h]
    \centering
    \begin{tabular}{l l c c c}
    \toprule
    Metric & Comparison & Mean Diff. & $t$ & $p_{\text{Holm}}$ \\
    \midrule
        \textbf{PRO} 
        & Position 1 vs 2 & \textbf{0.038} & \textbf{4.07} & \boldmath{$<.001^{***}$} \\
        & Position 1 vs 3 & \textbf{0.028} & \textbf{3.11} & \boldmath{$0.004^{**}$} \\
        & Position 2 vs 3 & -0.010 & -1.09 & 0.278 \\
    \midrule
    \textbf{NEUTRAL} 
        & Position 1 vs 2 & \textbf{0.027} & \textbf{3.72} & \boldmath{$<.001^{***}$} \\
        & Position 1 vs 3 & \textbf{0.028} & \textbf{3.74} & \boldmath{$<.001^{***}$} \\
        & Position 2 vs 3 & 0.001 & 0.13 & 0.898 \\
    \midrule
    \textbf{CON} 
        & Position 1 vs 2 & \textbf{0.024} & \textbf{2.88} & \boldmath{$0.014^{*}$} \\
        & Position 1 vs 3 & \textbf{0.021} & \textbf{2.46} & \boldmath{$0.030^{*}$} \\
        & Position 2 vs 3 & -0.003 & -0.37 & 0.712 \\
    \bottomrule
    \end{tabular}
    \caption{Post-hoc pairwise comparisons for BERTScore across the three stance conditions. 
    Holm correction applied within each stance (three comparisons per stance). 
    Significance codes: * $p<.05$, ** $p<.01$, *** $p<.001$.}
    \label{tab:posthoc_bertscore}
\end{table}
To examine whether the presentation order of input documents influences the similarity of the generated summaries, we conducted a series of one-way ANOVAs. For each metric–stance combination, the input position (1st, 2nd, 3rd) was treated as the independent variable and the similarity score as the dependent variable, testing the model configuration \textit{value}~$\sim$~\textit{position}. The resulting $p$-values were corrected across all tests using the Benjamini–Hochberg FDR procedure. Table~\ref{tab:anova} reports the $F$ statistics, FDR-corrected $p$-values, and corresponding $\eta^{2}$ effect sizes. The results show a clear positional effect for BERTScore: all three stances exhibit significant differences across input positions, with medium-sized effects for PRO and NEUTRAL ($p_{\text{corr}} < .001$, $\eta^{2} \approx .07$). The effect for CON is smaller but still statistically reliable ($p_{\text{corr}} = .030$, $\eta^{2} = .038$). 
In contrast, neither ROUGE-L nor SummaC show significant positional differences after correction.

To identify which input positions differed from one another specifically, we conducted post-hoc pairwise $t$-tests for all stance conditions that showed significant ANOVA effects. Holm-corrected results are presented in Table~\ref{tab:posthoc_bertscore}. Across all three stances, Position~1 consistently yielded higher BERTScore values than both Position~2 and Position~3, indicating that documents presented first contributed more to the model's generated summaries.

For PRO articles, Position~1 produced significantly higher similarity scores than Position~2 ($t=4.07$, $p_{\text{Holm}}<.001$) and Position~3 ($t=3.11$, $p_{\text{Holm}}=.004$), while the difference between Positions~2 and 3 was not significant. A similar pattern was observed for NEUTRAL articles, where Position~1 again surpassed Position~2 and Position~3 (both $p_{\text{Holm}}<.001$), with no detectable difference between the latter two positions. For CON articles, Position~1 also exceeded both Position~2 ($t=2.88$, $p_{\text{Holm}}=.014$) and Position~3 ($t=2.46$, $p_{\text{Holm}}=.030$), although effect sizes were smaller than for PRO and NEUTRAL. No significant difference was found between Positions~2 and 3. Together, these results reveal a consistent primacy effect across all stance categories, manifested as higher similarity scores when the PRO, NEUTRAL, or CON article appeared in the first position.

\section{Discussion}
To address RQ1, our results indicate a consistent trend across ROUGE-L and BERTScore: the generated summaries exhibit greater similarity to the first article in the input sequence. The absolute metric values observed in our study differ from those commonly reported in prior summarization research. The reason is that different from previous work that evaluates model outputs against human-written reference summaries \citep{10673650, Wang2023ElementawareSW}, our evaluation compares the generated summary directly with the full original articles. Because the summaries produced by the model average only ~220 words, while the source articles range from 300 to 1600 words, the strong length mismatch naturally lowers token-overlap and embedding-based similarity scores. Consequently, we observe the relative score on different input conditions instead of the absolute value of the metrics.
Another methodological constraint arises from the limited availability of metrics designed to assess similarity between a summary and its source documents. Existing evaluation frameworks are mainly reference-based, leaving few options for measuring source alignment. SummaC, which captures sentence-level entailment, yields low scores and shows minimal sensitivity to positional variation. Abstract summaries omit the details of source content, and entailment-based scoring penalizes unsupported or implicit statements; therefore, even if the model prioritizes information from the first article, it is unlikely to reuse its surface forms verbatim. As a result, SummaC cannot reward subtle content biases unless explicit supporting sentences are preserved.

Statistical tests show that the presentation order of input articles affects the generated summaries; specifically, LLMs show a consistent preference for the first input article, thereby shaping both the framing and the semantic similarity patterns. Prior work has documented position bias in LLMs when processing long contextual inputs \citep{Liu2023LostIT}. Our results extend this finding to the multi-document setting, demonstrating a clear primacy effect in information ranking scenarios.
BERTScore consistently reveals a primacy effect across all three stances. This pattern arises because primacy in LLM summarization primarily manifests as a shift in semantic framing where the model preferentially adopts the vocabulary and argumentative structure of the first-seen document. As a result, token-level contextualized embeddings produce higher similarity scores for Position~1.
By contrast, ROUGE-L, which is restricted to surface-level lexical overlap, remains low for abstractive summaries, and SummaC focuses on sentence-level factual entailment rather than framing similarity. Consequently, these metrics detect weaker or no primacy effects relative to BERTScore.
For all three stances, Position 1 differs significantly from both Position 2 and Position 3 under BERTScore, whereas the comparison between Position 2 and Position 3 is consistently non-significant. This pattern indicates that the primacy effect arises because the model assigns importance to the first article, rather than because later positions gradually receive less influence.

The consequences of this order effect extend to real-world information retrieval and user-facing AI systems. For example, the \texttt{AI Overviews} feature in Google Search, powered by Gemini models, may implicitly treat earlier-ranked articles as more important while overlooking those presented later. Such imbalanced attention can lead to skewed framing in the generated overviews and potentially influence users’ interpretation of the topic, particularly among users who place high trust in AI-generated summaries.

More importantly, when LLMs are embedded within agentic AI systems, where multiple agents coordinate to perform complex tasks, the function of LLMs becomes crucial for downstream decision-making. If primacy bias distorts these early steps, it can propagate through the entire agentic workflow. Therefore, addressing the primacy effect is essential for ensuring transparent, ethical, and trustworthy agentic behavior.\\
\textbf{Limitations}
This study has two main limitations. First, although the text length of each article triplet is controlled, we do not explicitly examine how sensitive the evaluation metrics are to document length. Because length may influence the absolute values of ROUGE-L, BERTScore, and SummaC, future work could categorize the dataset by length or conduct length-controlled analyses to better isolate positional effects. Second, as shown in our results, the absolute metric scores are relatively low. This reflects a broader limitation of reference-free evaluation metrics when applied to abstractive summarization, where direct comparison between long source documents and concise summaries remains challenging.\\
\textbf{Future Work}
Future work should extend the dataset beyond the abortion domain to other politically and socially controversial topics such as gun control, LGBTQ+ rights, and immigration to assess whether the observed order effects generalize across issue areas. In addition, evaluating multiple LLM families (e.g., GPT, Claude) would enable a more comprehensive examination of whether primacy effects persist across different model architectures. Finally, introducing additional evaluation metrics, including positional stance measures and social polarization indicators, could provide a more comprehensive characterization of order-sensitive behaviors in LLM-generated summaries.
\section{Conclusion}
We find that LLMs exhibit a clear primacy effect in multi-document news summarization; that is, the generated summaries are more aligned with the first-seen article, particularly at the semantic level. The findings extend prior evidence of position bias to multi-source summarization, showing that the order in which information is presented shapes the model’s output. Such position sensitivity poses risks for systems like \textit{AI Overviews} and agentic AI pipelines, where early-seen documents disproportionately influence downstream actions. Therefore, it is of significance to address primacy effects for developing more balanced, transparent, and trustworthy LLM-based information systems.

\clearpage
\bibliographystyle{plainnat}
\bibliography{ref}

\end{document}